\documentclass[10pt,twocolumn,letterpaper]{article}

\usepackage[pagenumbers]{cvpr} 
\usepackage{appendix}

\usepackage[dvipsnames]{xcolor}

\definecolor{cvprblue}{rgb}{0.21,0.49,0.74}
\usepackage[pagebackref,breaklinks,colorlinks,citecolor=cvprblue]{hyperref}
\usepackage{xcolor}
\usepackage{graphicx}
\usepackage{multirow}
\def\paperID{4245} 
\def\confName{CVPR}
\def\confYear{2024}

\title{Text-Guided 3D Face Synthesis - From Generation to Editing}

 \author{%
   Yunjie Wu$^{\dagger1}$ \and Yapeng Meng$^{\dagger1,2}$ \and Zhipeng Hu$^{\dagger1}$\and Lincheng Li$^{*1}$ \and
   Haoqian Wu$^{1}$ \and
   Kun Zhou$^{3}$ \;\;\; Weiwei Xu$^{3}$  \;\;\; Xin Yu$^{4}$
   \\
   $^1$Netease Fuxi AI Lab  \;\;\; $^2$Tsinghua University   \;\;\;$^3$Zhejiang University \;\;\;  $^4$University of Queensland
}

\definecolor{lightred}{RGB}{218, 125, 116} 
\definecolor{lightblue}{RGB}{157, 195, 250} 

\begin{document}
\twocolumn[{%
\renewcommand\twocolumn[1][]{#1}%
\maketitle

\vspace{-10mm}
\begin{center}
  \centering
   \includegraphics[width=0.989\linewidth]{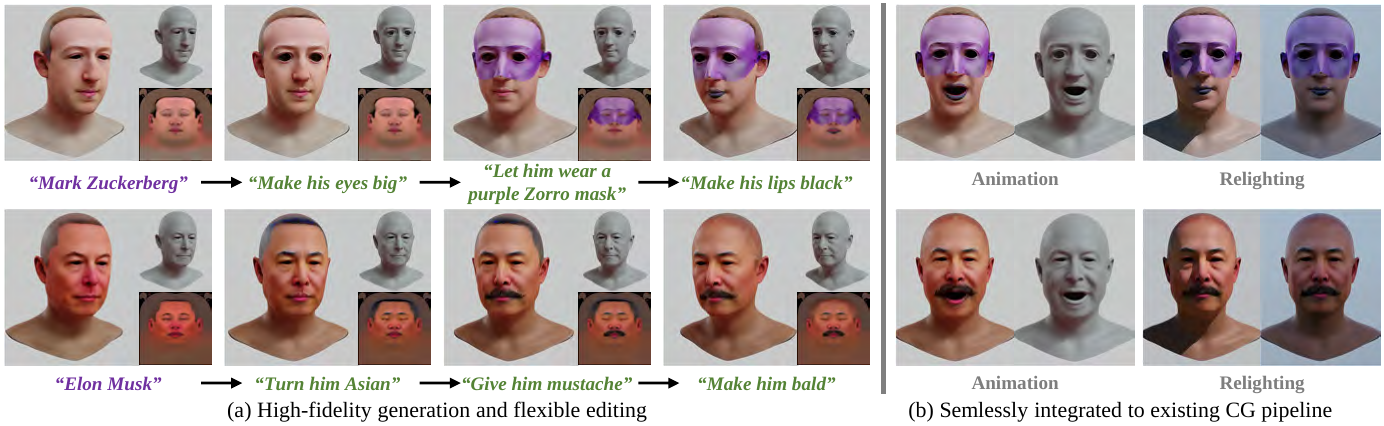}
   \vspace{-0.5em}
   \captionof{figure}{(a) Our approach enables the high-fidelity generation and flexible editing of 3D faces from textual input. It facilitates sequential editing for creating customized details in 3D faces. (b) The produced 3D faces can be seamlessly integrated into existing CG pipelines.}
   \label{fig:teaser}
\end{center}
}]

\def\thefootnote{$\dagger$}\footnotetext{Equal contribution}
\def\thefootnote{*}\footnotetext{Corresponding author}

\begin{abstract}
\vspace{-1.1em}
Text-guided 3D face synthesis has achieved remarkable results by leveraging text-to-image (T2I) diffusion models. However, most existing works focus solely on the direct generation, ignoring the editing, restricting them from synthesizing customized 3D faces through iterative adjustments. In this paper, we propose a unified text-guided framework from face generation to editing. In the generation stage, we propose a geometry-texture decoupled generation to mitigate the loss of geometric details caused by coupling. Besides, decoupling enables us to utilize the generated geometry as a condition for texture generation, yielding highly geometry-texture aligned results. We further employ a fine-tuned texture diffusion model to enhance texture quality in both RGB and YUV space. In the editing stage, we first employ a pre-trained diffusion model to update facial geometry or texture based on the texts. To enable sequential editing, we introduce a UV domain consistency preservation regularization, preventing unintentional changes to irrelevant facial attributes. Besides, we propose a self-guided consistency weight strategy to improve editing efficacy while preserving consistency. Through comprehensive experiments, we showcase our method's superiority in face synthesis. Project page: \url{https://faceg2e.github.io/}.

\end{abstract}


\section{Introduction}
\label{sec:intro}
Modeling 3D faces serves as a fundamental pillar for various emerging applications such as film making, video games, and AR/VR. 
Traditionally, the creation of detailed and intricate 3D human faces requires extensive time from highly skilled artists. With the development of deep learning, existing works \cite{zielonka2022towards,dey2022generating,wood20223d,dib2023s2f2} attempted to produce 3D faces from photos or videos with generative models. 
However, the diversity of the generation remains constrained primarily due to the limited scale of training data.
Fortunately, recent large-scale vision-language models (e.g., CLIP \cite{radford2021learning}, Stable Diffusion \cite{rombach2022high}) pave the way for generating diverse 3D content. Through the integration of these models, numerous text-to-3D works \cite{liu2022towards,jiang20233d,Xu_2023_CVPR,youwang2022clip,lin2023magic3d} can create 3D content in a zero-shot manner. 

Many studies have been conducted on text-to-3D face synthesis. They either utilize CLIP or employ score distillation sampling (SDS) on text-to-image (T2I) models to guide the 3D face synthesis. Some methods \cite{wang2023rodin,zhang2023avatarverse} employ neural fields to generate visually appealing but low-quality geometric 3D faces. Recently, Dreamface \cite{zhang2023dreamface} has demonstrated the potential for generating high-quality 3D face textures by leveraging SDS on facial textures, but their geometry is not fidelitous enough and they overlooked the subsequent face editing. A few works \cite{aneja2022clipface,han2023headsculpt,liao2023tada} enable text-guided face editing, allowing coarse-grained editing (e.g. overall style), but not fine-grained adjustments (e.g., lips color). Besides, the lack of design in precise editing control leads to unintended changes in their editing, preventing the synthesis of customized faces through sequential editing.

To address the aforementioned challenges, we present text-guided 3D face synthesis - from generation to editing, dubbed \textbf{FaceG2E}. We propose a progressive framework to generate the facial geometry and textures, and then perform accurate face editing sequentially controlled by text. To the best of our knowledge, this is the first attempt to edit a 3D face in a sequential manner. We propose two core components: (1) {{Geometry-texture decoupled generation}} and (2) {{Self-guided consistency preserved editing}}.

To be specific, our proposed \textit{\textbf{Geometry-texture decoupled generation}} generates the facial geometry and texture in two separate phases. By incorporating texture-less rendering in conjunction with SDS, we induce the T2I model to provide geometric-related priors, inciting details (e.g., wrinkles, lip shape) in the generated geometry. Building upon the generated geometry, we leverage ControlNet to force the SDS to be aware of the geometry, ensuring precise geometry-texture alignment. Additionally, we fine-tune a texture diffusion model that incorporates both RGB and YUV color spaces to compute SDS in the texture domain, enhancing the quality of the generated textures.

The newly developed \textit{\textbf{Self-guided consistency preserved editing}} enables one to follow the texts, performing efficient editing in specific facial attributes without causing other unintended changes. 
Here, we first employ a pre-trained image-edit diffusion model to update the facial geometry or texture. Then we introduce a UV domain consistency preservation regularization to prevent unexpected changes in faces, enabling sequential editing. To avoid the degradation of editing effects caused by the regularization, we further propose a self-guided consistency weighting strategy. It adaptively determines the regularization weight for each facial region by projecting the cross-attention scores of the T2I model to the UV domain.
As shown in Fig. \ref{fig:teaser}, our method can generate high-fidelity 3D facial geometry and textures while allowing fine-grained face editing. With the proposed components, we achieve better visual and quantitative results compared to other SOTA methods, as demonstrated in Sec. \ref{sec:experiment}.
In summary, our contributions are:
\begin{itemize}
    \item We propose FaceG2E, facilitating a full pipeline of text-guided 3D face synthesis, from generation to editing. User surveys confirm that our synthesized 3D faces are significantly preferable than other SOTA methods. 
    \item We propose the geometry-texture decoupled generation, producing faces with high-fidelity geometry and texture.
    \item We design the self-guided consistency preservation, enabling the accurate editing of 3D faces. Leveraging precise editing control, our method showcases some novel editing applications, such as sequential and geometry-texture separate editing.
\end{itemize}

\section{Related Work}
\label{sec:relat}

\noindent\textbf{Text-to-Image generation.} Recent advancements in visual-language models \cite{radford2021learning} and diffusion models \cite{ho2020denoising,song2020denoising,dhariwal2021diffusion} have greatly improved text-to-image generation \cite{ramesh2022hierarchical,saharia2022photorealistic,rombach2022high,balaji2022ediffi}. These methods, trained on large-scale image-text datasets \cite{schuhmann2022laion,schuhmann2021laion}, can synthesize realistic and complex images from text descriptions. Subsequent studies have made further efforts to introduce additional generation process controls \cite{xie2023boxdiff,zhang2023adding,huang2023composer}, fine-tuning the pre-trained models for specific scenarios \cite{ruiz2023dreambooth,hu2021lora,gal2022image}, and enabling image editing capabilities \cite{hertz2022prompt,brooks2023instructpix2pix,kawar2023imagic}. However, generating high-quality and faithful 3D assets, such as 3D human faces, from textual input still poses an open and challenging problem.

\noindent\textbf{Text-to-3D generation.} With the success of text-to-image generation in recent years, text-to-3D generation has attracted significant attention from the community. Early approaches \cite{xu2023dream3d,hong2022avatarclip,michel2022text2mesh,sanghi2022clip,jain2022zero} utilize mesh or implicit neural fields to represent 3D content, and optimized the CLIP metrics between the 2D rendering and text prompts. However, the quality of generated 3D contents is relatively low.

Recently, DreamFusion \cite{poole2022dreamfusion} has achieved impressive results by using a score distillation sampling (SDS) within the powerful text-to-image diffusion model \cite{saharia2022photorealistic}. Subsequent works further enhance DreamFusion by reducing generation time \cite{lin2023magic3d}, improving surface material representation \cite{chen2023fantasia3d}, and introducing refined sampling strategies \cite{huang2023dreamtime}. However, the text-guided generation of high-fidelity and intricate 3D faces remains challenging. Building upon DreamFusion, we carefully design the form of score distillation by exploiting various diffusion models at each stage, resulting in high-fidelity and editable 3D faces.

\begin{figure*}[t]
  \centering
   \includegraphics[width=0.93\linewidth]{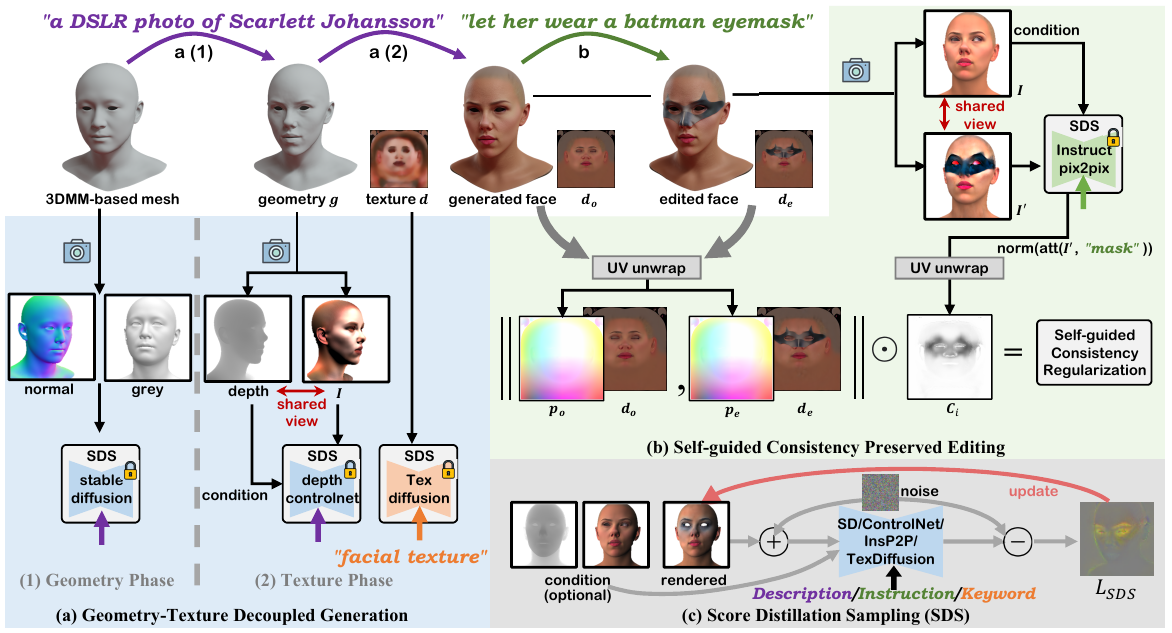}
   \vspace{-0.5em}
   \caption{Overview of FaceG2E. (a) Geometry-texture decoupled generation, including a geometry phase and a texture phase. (b) Self-guided consistency preserved editing, in which we utilize the built-in cross-attention to obtain the editing-relevant regions and unwrap them to UV space. Then we penalize inconsistencies in the irrelevant regions. (c) Our method exploits multiple score distillation sampling.}
   \label{fig:overview}
\end{figure*}

\noindent\textbf{Text-to-3D face synthesis.} 
Recently, there have been attempts to generate 3D faces from text. Describe3D \cite{wu2023high} and Rodin \cite{wang2023rodin} propose to learn the mapping from text to 3D faces on pairs of text-face data. They solely employ the mapping network trained on appearance descriptions to generate faces, and thus fail to generalize to out-of-domain texts (e.g., celebrities or characters). On the contrary, our method can generalize well to these texts and synthesize various 3D faces.

Other works \cite{zhang2023dreamface,han2023headsculpt,liao2023tada,huang2023humannorm,jiang2023avatarcraft} employ SDS on the pre-trained T2I models. Dreamface \cite{zhang2023dreamface} utilizes CLIP to select facial geometry from candidates. Then they perform the SDS with a texture diffusion network to generate facial textures. Headsculpt \cite{han2023headsculpt} employs Stable Diffusion \cite{rombach2022high} and InstructPix2Pix \cite{brooks2023instructpix2pix} for computing the SDS, and relies on the mixture of SDS gradients for constraining the editing process. These approaches can perform not only generation but also simple editing. However, they still lack the design in precise editing control, and unintended changes in the editing results often occur. This prevents them from synthesizing highly customized 3D faces via sequential editing. On the contrary, our approach facilitates accurate editing of 3D faces, supporting sequential editing.



\section{Methodology}
\label{sec:method}
FaceG2E is a progressive text-to-3D approach that first generates a high-fidelity 3D human face and then performs fine-grained face editing. As illustrated in Fig. \ref{fig:overview}, our method has two main stages: (a) Geometry-texture decoupled generation, and (b) Self-guided consistency preserved editing. In Sec. \ref{sec:3-1}, we introduce some preliminaries that form the fundamental basis of our approach. In Sec. \ref{sec:3-2} and Sec. \ref{sec:3-3}, we present the generation and editing stages.

\subsection{Preliminaries}
\label{sec:3-1}
\noindent\textbf{Score distillation sampling} has been proposed in DreamFusion \cite{poole2022dreamfusion} for text-to-3D generation. It utilizes a pre-trained 2D diffusion model $\phi$ with a denoising function $\epsilon_{\phi}\left(z_{t};y,t\right)$ to optimize 3D parameters $\theta$. SDS renders an image $I = R(\theta)$ and embeds $I$ with an encoder $\mathcal{E}(\cdot)$, achieving image latent $z$. Then it injects a noise $\epsilon$ into $z$, resulting in a noisy latent code $z_{t}$. It takes the difference between the predicted and added noise as the gradient:
\begin{equation}
    {\nabla_{\theta} \mathcal{L}_{\mathrm{SDS}}(I)=\mathbb{E}_{t, \epsilon}\left[w(t)\left(\epsilon_{\phi}\left(z_{t} ; y, t\right)-\epsilon\right) \frac{\partial z}{\partial I} \frac{\partial I}{\partial \theta}\right]},
\end{equation}


\noindent where $w(t)$ is a time-dependent weight function and $y$ is the embedding of input text.

\noindent\textbf{Facial Geometry and Texture} \noindent is represented with parameters $ \theta=(\beta,u)$ in FaceG2E. $\beta$ denotes the identity coefficient from the parametric 3D face model HIFI3D \cite{HIFI3D}, and $u$ denotes a image latent code for facial texture. The geometry $g$ can be achieved by the blendshape function $\mathbf{M}(\cdot)$:
\begin{equation}
g=\mathbf{M}(\beta)=T+\sum_{i} \beta_{i} \mathrm{S}_{i},
\end{equation}
where $T$ is the mean face and $\mathrm{S}$ is the vertices offset basis. As to the texture, the facial texture map $d$ is synthesized with a decoder: $d=\mathcal{D}(u)$. We take
the decoder from VAE of Stable Diffusion \cite{rombach2022high} as $\mathcal{D}(\cdot)$.

\subsection{Geometry-Texture Decoupled Generation}
\label{sec:3-2}
The first stage of FaceG2E is the geometry-texture decoupled generation, which generates facial geometry and texture from the textual input. Many existing works have attempted to generate geometry and texture simultaneously in a single optimization process, while we instead decouple the generation into two distinct phases: the geometry phase and the texture phase. The decoupling provides two advantages: 1) It helps enhance geometric details in the generated faces. 2) It improves geometry-texture alignment by exploiting the generated geometry to guide the texture generation.

\noindent\textbf{Geometry Phase.} 
An ideal generated geometry should possess both high quality (e.g., no surface distortions) and a good alignment with the input text. The employed facial 3D morphable model provides strong priors to ensure the quality of generated geometry. As to the alignment with the input text, we utilize SDS on the network $\phi_{sd}$ of Stable Diffusion \cite{rombach2022high} to guide the geometry generation.

Previous works \cite{liao2023tada,jiang2023avatarcraft,zhang2023avatarverse} optimize geometry and texture simultaneously. We observe this could lead to the loss of geometric details, as certain geometric information may be encapsulated within the texture representation. Therefore, we aim to enhance the SDS to provide more geometry-centric information in the geometry phase.
To this end, we render the geometry $g$ with texture-less rendering $\tilde{I}=\tilde{R}(g)$, e.g., surface normal shading or diffuse shading with constant grey color. The texture-less shading attributes all image details solely to geometry, thereby allowing the SDS to focus on geometry-centric information. The \textit{\textbf{geometry-centric SDS}} loss is defined as:
\begin{equation}
    {\nabla_{\beta} \mathcal{L}_{\mathrm{geo}}\!=\!\mathbb{E}_{t, \epsilon}\!\!\left[w(t)\left(\epsilon_{\phi_{sd}}\left(z_{t} ; y, t\right)-\epsilon\right) \frac{\partial z_{t}}{\partial \tilde{I}} \frac{\partial \tilde{I}}{\partial g}
    \frac{\partial g}{\partial \beta}\right]}.
\end{equation}

\noindent\textbf{Texture Phase.} Many works \cite{zhang2023dreamface,liao2023tada} demonstrate that texture can be generated by minimizing the SDS loss. However, directly optimizing the standard SDS loss could lead to geometry-texture misalignment issues, as shown in Fig .\ref{fig:ablation-generation}. To address this problem, we propose the \textit{\textbf{geometry-aware texture content SDS}} (GaSDS). We resort to the ControlNet \cite{zhang2023adding} to endow the SDS with awareness of generated geometry, thereby inducing it to uphold geometry-texture alignment. Specifically, we render $g$ into a depth map $e$. Then we equip the SDS with the depth-ControlNet $\phi_{dc}$, and take $e$ as a condition, formulating the GaSDS:
\begin{equation}
\nabla_{u}\mathcal{L}_{\mathrm{tex}}^{ga} = \mathbb{E}_{t, \epsilon}\left[w(t)\left(\epsilon_{\phi_{dc}}\left(z_{t} ; e, y, t\right)-\epsilon\right) \frac{\partial z_{t}}{\partial {I}} \frac{\partial {I}}{\partial d}
    \frac{\partial d}{\partial u}\right].
\end{equation}

With the proposed GaSDS, the issue of geometric misalignment is addressed. However, artifacts such as local color distortion or uneven brightness persist in the textures. This is because the T2I model lacks priors of textures, which hinders the synthesis of high-quality texture details.
\begin{figure}[t]
  \centering
   \includegraphics[width=0.93\linewidth]{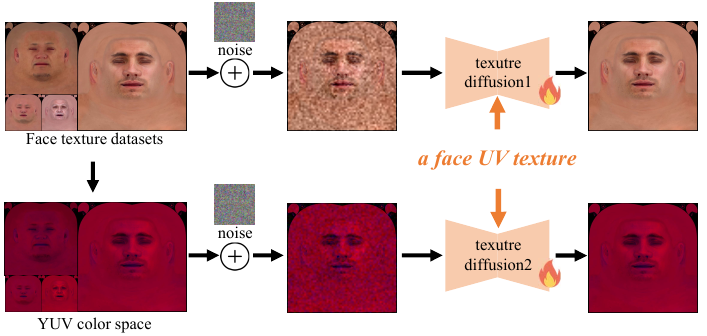}
   \vspace{-0.5em}
   \caption{Training the texture diffusion model is performed on the collected facial textures in both RGB and YUV color space.}
   \label{fig:texture-diffusion}
\end{figure}

Hence we propose \textit{\textbf{texture prior SDS}} to introduce such priors of textures. Inspired by DreamFace \cite{zhang2023dreamface}, we train a diffusion model $\phi_{td1}$ on texture data to estimate the texture distribution for providing the prior. Our training dataset contains 500 textures, including processed scanning data and selected synthesized data \cite{bai2022ffhq}. Different from DreamFace, which uses labeled text in training, we employ a fixed text keyword (e.g., `\textit{facial texture}') for all textures. Because the objective of $\phi_{td1}$ is to model the distribution of textures as a prior, the texture-text alignment is not necessary. We additionally train another $\phi_{td2}$ on the YUV color spaces to promote uniform brightness, as shown in Fig \ref{fig:texture-diffusion}. We fine-tune both $\phi_{td1}$ and $\phi_{td2}$ on Stable Diffusion. The texture prior SDS is formulated with the trained $\phi_{td1}$ and $\phi_{td2}$ as:
\begin{equation}
\begin{split}
&\nabla_{u}\mathcal{L}_{\mathrm{tex}}^{pr} = \mathcal{L}_{\mathrm{tex}}^{rgb} + \lambda_{yuv}\mathcal{L}_{\mathrm{tex}}^{yuv},\\
\mathrm{L}_{\mathrm{tex}}^{rgb} =\mathbb{E}_{t, \epsilon}&\left[w(t)\left(\epsilon_{\phi_{td1}}\left(z_{t}^{d} ; y^\ast, t\right)-\epsilon\right) \frac{\partial z_{t}^{d}}{\partial {d}} \frac{\partial {d}}{\partial u}\right],\\
\mathcal{L}_{\mathrm{tex}}^{yuv} = \mathbb{E}_{t, \epsilon}&\left[w(t)\left(\epsilon_{\phi_{td2}}\left(z_{t}^{d'} ; y^\ast, t\right)-\epsilon\right) \frac{\partial z_{t}^{d'}}{\partial {d}} \frac{\partial {d}}{\partial u}\right],
\end{split}
\end{equation}
where $z_{t}^{d}$ and $z_{t}^{d'}$ denote the noisy latent codes of the texture $d$ and the converted YUV texture $d'$. The $y^\ast$ is the text embedding of the fixed text keyword. We combine the $\mathcal{L}_{\mathrm{tex}}^{ga}$ and $\mathcal{L}_{\mathrm{tex}}^{pr}$ as our final texture generation loss:
\begin{equation}
\mathcal{L}_{\mathrm{tex}} = \mathcal{L}_{\mathrm{tex}}^{ga} + \lambda_{pr}\mathcal{L}_{\mathrm{tex}}^{pr},
\end{equation}
where $\lambda_{pr}$ is a weight to balance the gradient from $\mathcal{L}_{\mathrm{tex}}^{pr}$.

\subsection{Self-guided Consistency Preserved Editing}
\label{sec:3-3} To attain the capability of following editing instructions instead of generation prompts, a simple idea is to take the text-guided image editing model InstructPix2Pix \cite{brooks2023instructpix2pix} $\phi_{ip2p}$ as a substitute for Stable Diffusion to form the SDS:
\begin{equation}
\nabla_{\beta,u}\mathcal{L}_{\mathrm{edit}} = \mathbb{E}_{t, \epsilon}\left[w(t)\left(\epsilon_{\phi_{ip2p}}\left(z_{t}'; z_{t}, y^\ast, t\right)-\epsilon\right) \frac{\partial z_{t}'}{\partial \beta, \partial u}\right],
\label{eqi:face-editing}
\end{equation}
where $z_{t}'$ denotes the latent for the rendering of the edited face, and the original face is embedded to $z_{t}$ as an extra conditional input, following the setting of InstructPix2Pix. 

Note that our geometry and texture are represented by separate parameters $\beta$ and $u$, so it is possible to independently optimize one of them, enabling separate editing of geometry and texture. Besides, when editing the texture, we integrate the $\mathcal{L}_{\mathrm{tex}}^{pr}$ to maintain the structural rationality of textures.

\noindent\textbf{Self-guided Consistency Weight.}
The editing SDS in Eq. \ref{eqi:face-editing} enables effective facial editing, while fine-grained editing control still remains challenging, e.g., unpredictable and undesired variations may occur in the results, shown as Fig. \ref{fig:ablation-scp}. This hinders sequential editing, as earlier edits can be unintentionally disrupted by subsequent ones. Therefore, consistency between the faces before and after the editing should be encouraged.

However, the consistency between faces during editing and the noticeability of editing effects, are somewhat contradictory. Imagine a specific pixel in texture, encouraging consistency inclines the pixel towards being the same as the original pixel, while the editing may require it to take on a completely different value to achieve the desired effect.

A key observation in addressing this issue is that the weight of consistency should vary in different regions: For regions associated with editing instructions, a lower level of consistency should be maintained as we prioritize the editing effects. Conversely, for irrelevant regions, a higher level of consistency should be ensured. For instance, given the instruction ``\textit{let her wear a Batman eyemask}'', we desire the eyemask effect near the eyes region while keeping the rest of the face unchanged. 

\begin{figure}[t]
  \centering
   \includegraphics[width=0.93\linewidth]{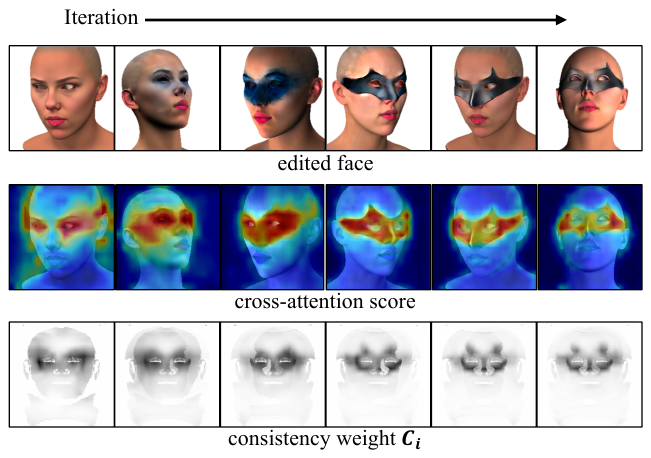}
   \vspace{-1em}
   \caption{Visualization of the edited face, the cross-attention score for token ``mask'' and the consistency weight $C_{i}$ during iterations in editing. Note the viewpoints vary due to random sampling in iterations.}
   \label{fig:scp}
\end{figure}

To locate the relevant region for editing instructions, we propose a self-guided consistency weight strategy in the UV domain. We utilize the built-in cross-attention of the InstructPix2Pix model itself. The attention scores introduce the association between different image regions and specific textual tokens. An example of the consistency weight is shown in Fig \ref{fig:scp}. We first select a region-indicating token $T^\ast$ in the instruction, such as ``\textit{mask}''. At each iteration $i$, we extract the attention scores between the rendered image $I$ of the editing and the token $T^\ast$. The scores are normalized and unwrapped to the UV domain based on the current viewpoint, and then we compute temporal consistency weight $\tilde{C_{i}}$ from the unwrapped scores:
\begin{equation}
\tilde{C_{i}}=1-\left(\operatorname{proj}\left(\operatorname{norm}\left(\operatorname{att}\left(I', T^{*}\right)\right)\right)\right)^{2},
\end{equation}
where $\operatorname{att}(\cdot,\cdot)$ denotes the cross-attention operation to predict the attention scores, the $\operatorname{norm}(\cdot)$ denotes the normalization operation, and the $\operatorname{proj}$ denotes the unwrapping projection from image to UV domain. As $\tilde{C_{i}}$ is related to the viewpoint, we establish a unified consistency weight ${C_{i}}$ to fuse $\tilde{C_{i}}$ from different viewpoints. The initial state of ${C_{i}}$ is a matrix of all `one', indicating the highest level of consistency applied to all regions. The updating of ${C_{i}}$ at each step is informed by the $\tilde{C_{i}}$. Specifically, we select the regions where the values in $\tilde{C_{i}}$ are lower than ${C_{i}}$ to be updated. Then we employ a moving average strategy to get the $C_{i}$:
\begin{equation}
\begin{split}
C_{i}&=C_{i-1} * w+\tilde{C}_{i} *(1-w),
\end{split}
\end{equation}
where $w$ is a fixed moving average factor. We take the $C_{i}$ as a weight to perform region-specific consistency.

\noindent\textbf{Consistency Preservation Regularization.}
With the consistency weight $C_{i}$ in hand, we propose a region-specific consistency preservation regularization in the UV domain to encourage consistency between faces before and after editing in both texture and geometry:
\begin{equation}
\begin{split}
\mathcal{L}_{\mathrm{reg}}^{tex}&=\left\|(d_{o}-d_{e})\odot C_{i}\right\|_{2}^{2},\\
\mathcal{L}_{\mathrm{reg}}^{geo}&=\left\|(p_{o}-p_{e})\odot C_{i}\right\|_{2}^{2},
\end{split}
\end{equation}
where $d_{o}$, $d_{e}$ denote the texture before and after the editing, $p_{o}$, $p_{e}$ denote the vertices position map unwrapped from the facial geometry before and after the editing, and $\odot$ denotes the Hadamard product.

With the consistency preservation regularization, we propose the final loss for our self-guided consistency preserved editing as:
\begin{equation}
L_{finalEdit}=L_{\text {edit}}+\lambda_{reg}
L_{\text {reg}},
\end{equation}
where $\lambda_{reg}$ is the balance weight.

\section{Experiments}
\label{sec:experiment}
\subsection{Implementation Details}
Our implementation is built upon Huggingface Diffusers \cite{von2022diffusers}. We use \textit{stable-diffusion} \cite{sd15} checkpoint for geometry generation, and \textit{sd-controlnet-depth} \cite{controlnet} for texture generation. We utilize the official \textit{instruct-pix2pix} \cite{ip2p} in face editing. The RGB and YUV texture diffusion models are both fine-tuned on the \textit{stable-diffusion} checkpoint. We utilize NVdiffrast \cite{Laine2020diffrast} for differentiable rendering. Adam \cite{kingma2014adam} optimizer with a fixed learning rate of 0.05 is employed. The generation and editing for geometry/texture require 200/400 iterations, respectively. It takes about 4 minutes to generate or edit a face on a single NVIDIA A30 GPU. We refer readers to the supplementary material for more implementation details.

\begin{figure*}[t]
  \centering
   \includegraphics[width=.992\linewidth]{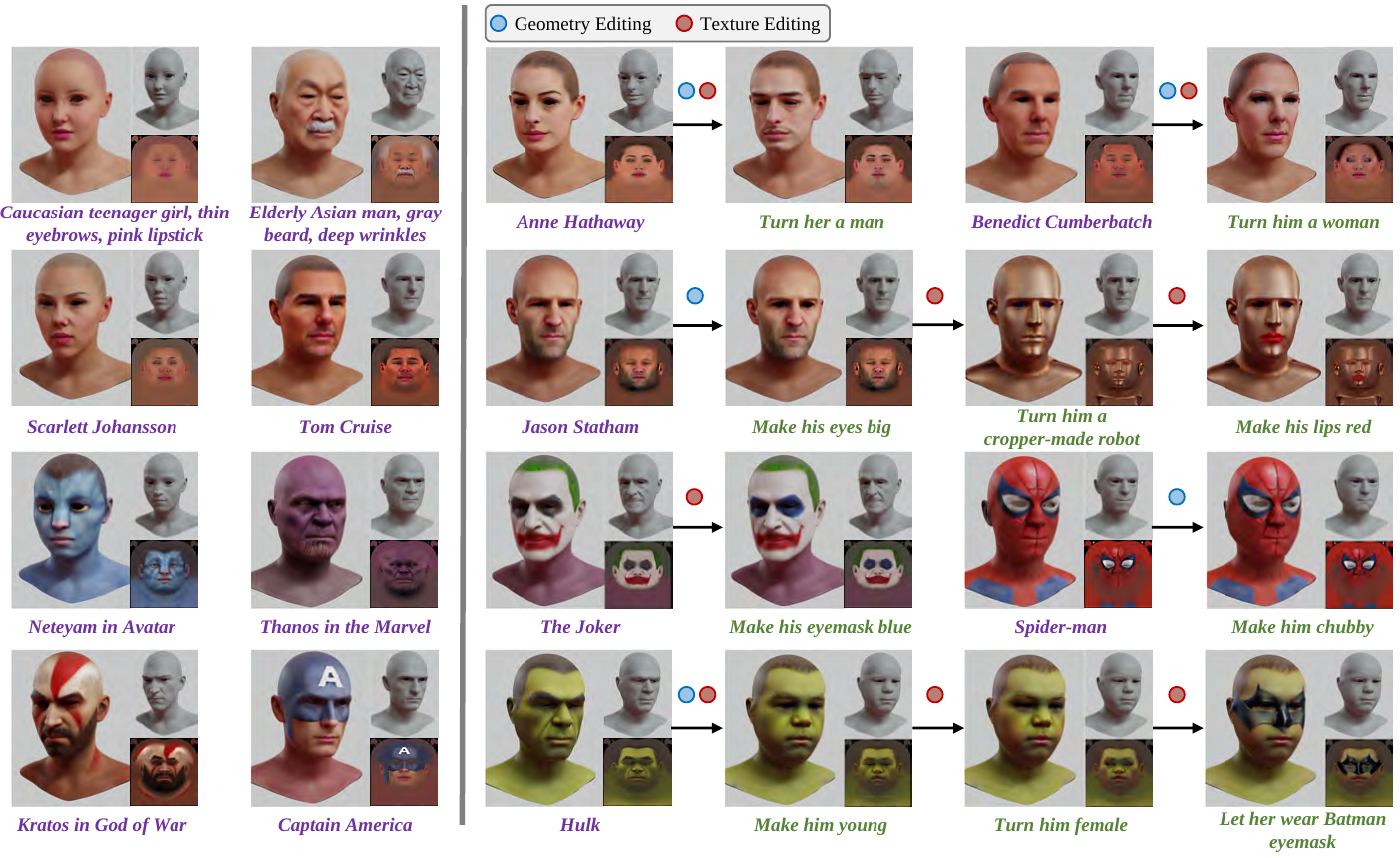}
   \caption{FaceG2E enables the generation of highly realistic and diverse 3D faces (on the left), as well as provides flexible editing capabilities for these faces (on the right). Through sequential editing, FaceG2E achieves the synthesis of highly customized 3D faces, such as `A female child Hulk wearing a Batman mask'. Additionally, independent editing is available for geometry and texture modification.}
   \label{fig:synthesis_result}
\end{figure*}
\subsection{Synthesis Results}
We showcase some synthesized 3D faces in Fig. \ref{fig:teaser} and Fig. \ref{fig:synthesis_result}. As depicted in the figures, FaceG2E demonstrates exceptional capabilities in generating a wide range of visually diverse and remarkably lifelike faces, including notable celebrities and iconic film characters. Furthermore, it enables flexible editing operations, such as independent manipulation of geometry and texture, as well as sequential editing. Notably, our synthesized faces can be integrated into existing CG pipelines, enabling animation and relighting applications, as exemplified in Fig. \ref{fig:teaser}. More animation and relighting results are in the supplementary material.

\subsection{Comparison with the state-of-the-art}
We compare some state-of-the-art methods for text-guided 3D face generation and editing, including Describe3D \cite{wu2023high}, DreamFace \cite{zhang2023dreamface} and TADA \cite{liao2023tada}. Comparisons with some other methods are contained in the supplementary material.

\subsubsection{Qualitative Comparison}
\begin{figure}[t]
  \centering
   \includegraphics[width=\linewidth]{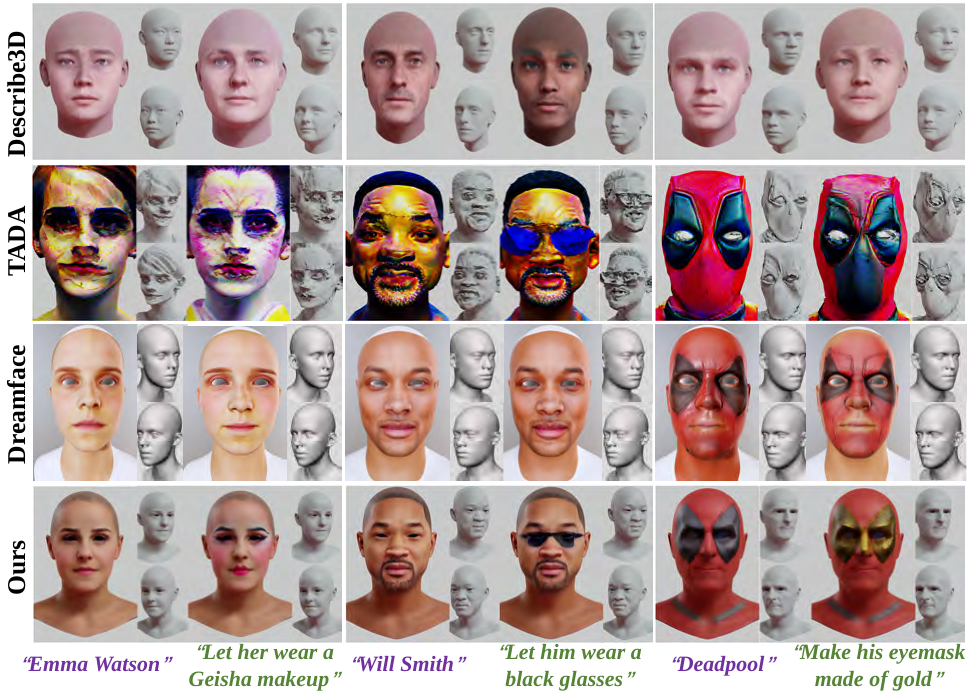}
   \caption{The comparison on text-guided 3D face synthesis. We present both the generation and editing results of each method.}
   \label{fig:qualitative comparison}
\end{figure}
\begin{figure}[t]
  \centering
   \includegraphics[width=\linewidth]{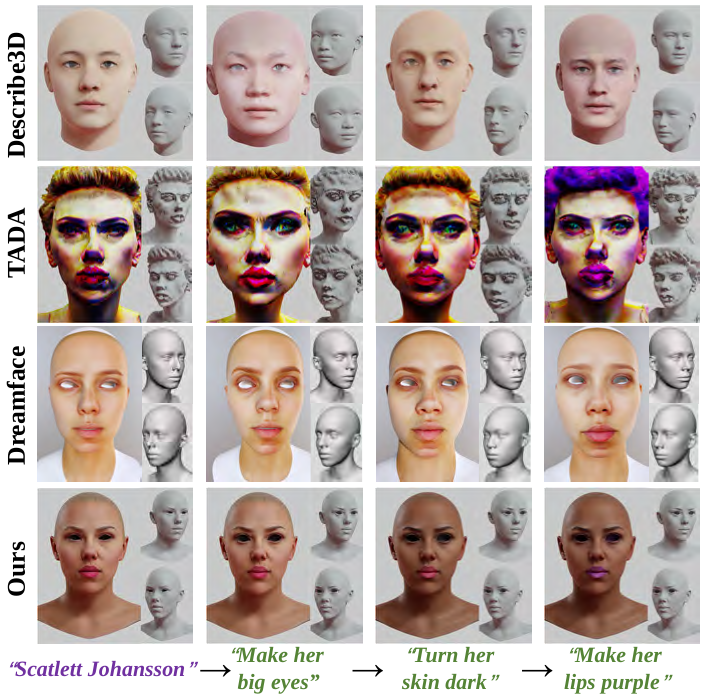}
   \caption{The comparison on sequential face editing.}
   \label{fig:sequential editing}
\end{figure}
The qualitative results are presented in Fig. \ref{fig:qualitative comparison}. We can observe that: (1) Describe3D struggles to generate 3D faces following provided texts due to its limited training data and inability to generalize beyond the training set. 
(2) TADA produces visually acceptable results but exhibits shortcomings in (i) generating high-quality geometry (e.g., evident geometric distortion in its outputs), and (ii) accurately following editing instructions (e.g., erroneously changing black glasses to blue in case 2).
(3) Dreamface can generate realistic faces but lacks editing capabilities.  Moreover, its geometry fidelity is insufficient, hindering the correlation between the text and texture-less geometry. In comparison, our method is superior in both generated geometry and texture and allows for accurate and flexible face editing. 

We further provide a comparison of sequential editing in Fig. \ref{fig:sequential editing}. Clearly, the editing outcomes of Describe3D and Dreamface in each round lack prominence. Although TADA performs well with single-round editing instructions, it struggles in sequence editing due to unintended changes that impact the preceding editing effects influenced by subsequent edits. For instance, in the last round, TADA mistakenly turns the skin purple. In contrast, our FaceG2E benefits from the proposed self-guided consistency preservation, allowing for precise sequence editing.

\subsubsection{Quantitative Comparison}
\begin{table}
\centering
\resizebox{.92\linewidth}{!}{
\begin{tabular}{c|cc|cc}
\hline
\multicolumn{1}{c|}{\multirow{2}{*}{Method}} & \multicolumn{2}{c|}{Generation} & \multicolumn{2}{c}{Editing}                                             \\ \cline{2-5} 
\multicolumn{1}{c|}{}                        & Score ↑ & Ranking-1 ↑ & \multicolumn{1}{c}{Score ↑} & \multicolumn{1}{c}{Ranking-1 ↑} \\ \hline
Describe3D \cite{wu2023high}                                  & 29.81        & 0\%              & 28.83                            & 0\%                                  \\
Dreamface \cite{zhang2023dreamface}                                   & 33.22        & 10\%             & 33.14                            & 10\%                                 \\
TADA \cite{liao2023tada}                                        & 34.85        & 10\%             & 33.73                            & 20\%                                 \\
Ours                                         & \textbf{36.95}        & \textbf{80}\%             & \textbf{35.50}                          & \textbf{70}\%                                 \\ \hline
\end{tabular}}
\caption{The CLIP evaluation results on the synthesized 3D faces.}
\label{table:clip-score}
\vspace{-0.2cm}
\end{table}

We quantitatively compare the fidelity of synthesized faces to text descriptions using the CLIP evaluation. We provide a total of 20 prompts, evenly split between generation and editing tasks, to all methods for face synthesis. All results are rendered with the same pipeline, except DreamFace, which takes its own rendering in the web demo \cite{dreamfacedemo}. A fixed prefix `a realistic 3D face model of ' is employed for all methods when calculating the CLIP score. We report the CLIP Score \cite{sanghi2023clip} and Ranking-1 in Tab. \ref{table:clip-score}. CLIP Ranking-1 calculates the ratio of a method's created faces ranked as top-1 among all methods. The results validate the superior performance of our method over other SOTA methods.

\subsubsection{User Study}
\begin{figure}[t]
  \centering
   \includegraphics[width=\linewidth]{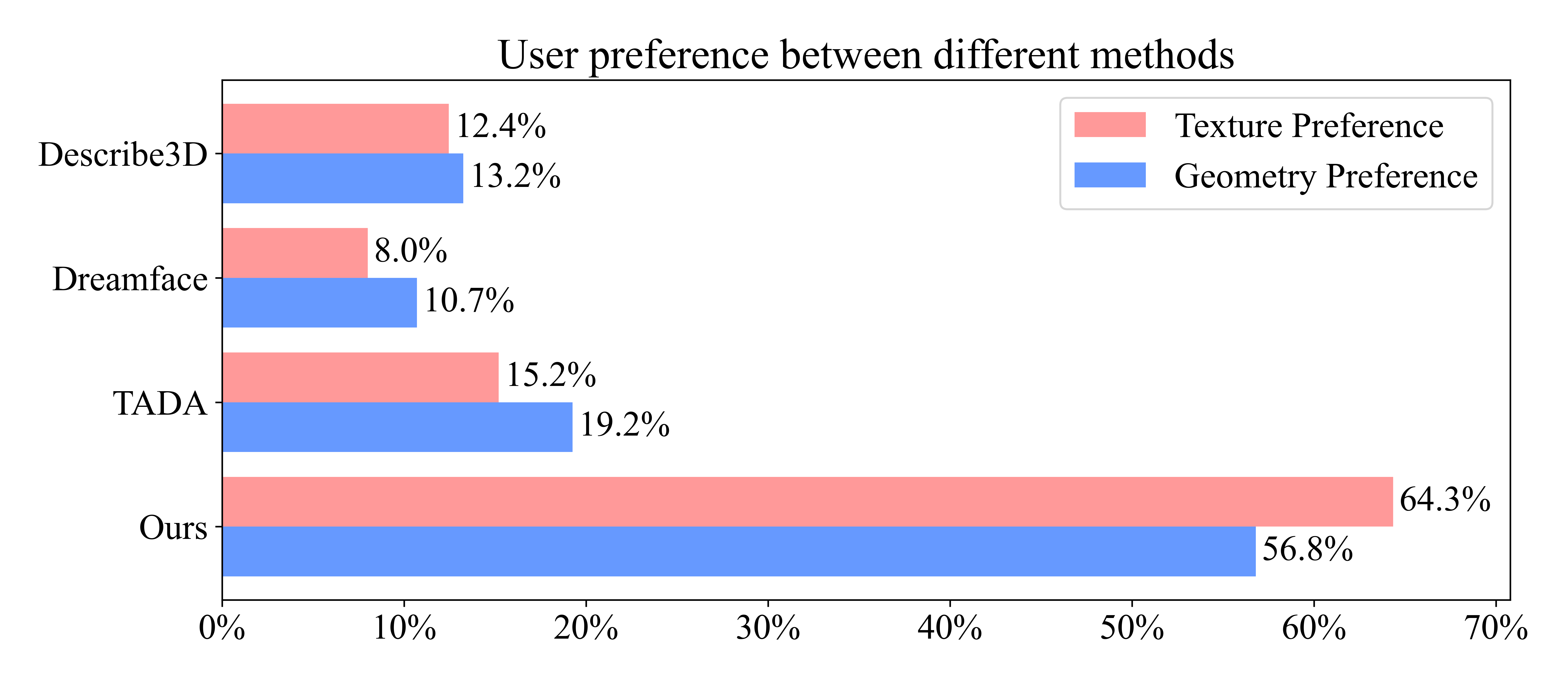}
   \caption{Quantitative results of user study. Our results are more favored by the participants compared to the other methods.}
   \label{fig:user-study}
   \vspace{-0.5em}
\end{figure}
We perform a comparative user study involving 100 participants to evaluate our method against state-of-the-art (SOTA) approaches. Participants are presented with 10 face generation examples and 10 face editing examples, and are asked to select the best method for each example based on specific criteria. The results, depicted in Fig. \ref{fig:user-study}, unequivocally show that our method surpasses all others in terms of both geometry and texture preference.

\subsection{Ablation Study}
Here we present some ablation studies. Extra studies based on user surveys are provided in the supplementary material.
\vspace{-1.em}
\begin{figure*}[t]
    \centering
    \includegraphics[width=\linewidth]{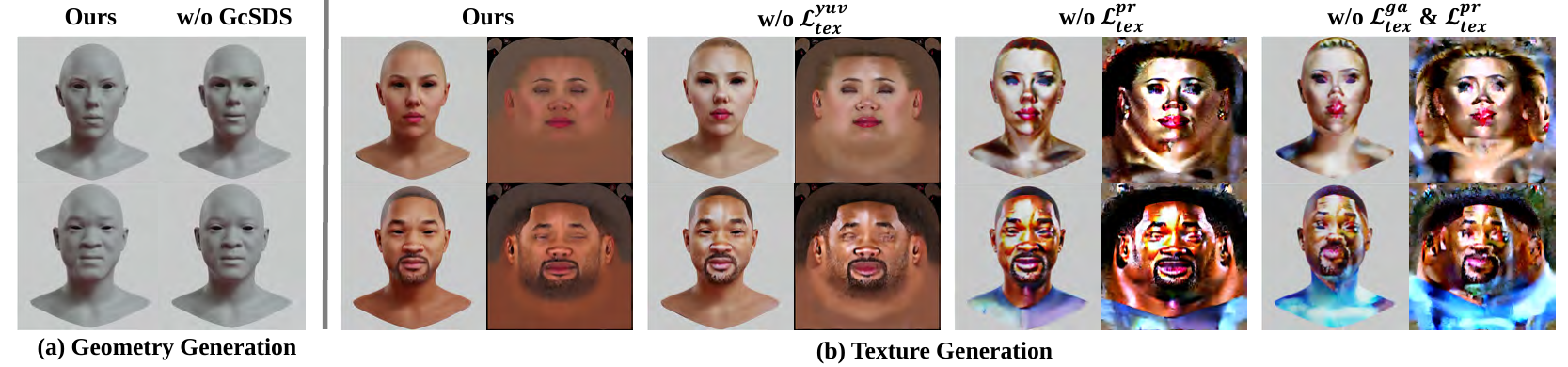}
    \captionsetup{width=\linewidth}
    \caption{The ablation study of our geometry-texture decoupled generation. The input texts are `Scarlett Johansson' and `Will Smith'.}
    \label{fig:ablation-generation}

\end{figure*}
\begin{figure}[t]
    \centering
    \includegraphics[width=\linewidth]{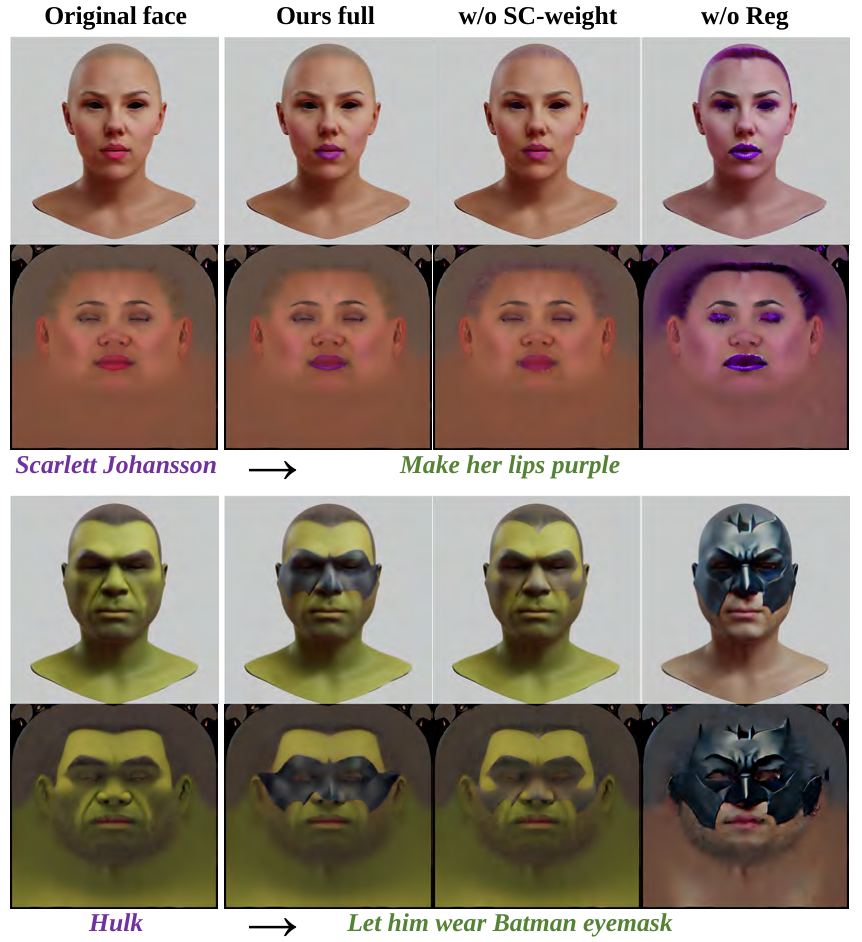}
    \captionsetup{width=\linewidth}
    \caption{Analysis of the proposed self-guided consistency preservation (SCP) in 3D face editing.}
    \label{fig:ablation-scp}
\vspace{-0.18cm}

\end{figure}
\subsubsection{Effectiveness of GDG}
To evaluate the effectiveness of geometry-texture decoupled generation (GDG), we conduct the following studies.

\noindent\textbf{Geometry-centric SDS (GcSDS)}. In Fig. \ref{fig:ablation-generation}(a), we conduct an ablation study to assess the impact of the proposed GcSDS. We propose a variation that takes standard textured rendering as input for SDS and simultaneously optimizes both geometry and texture variables. The results reveal that without employing the GcSDS, there is a tendency to generate relatively planar meshes, which lack geometric details such as facial wrinkles. We attribute this deficiency to the misrepresentation of geometric details by textures.

\noindent\textbf{Geometry-aligned texture content SDS (GaSDS)}. In Columns 3 and 4 of Fig. \ref{fig:ablation-generation}(b), we evaluate the effectiveness of GaSDS. We replace the depth-ControlNet in GaSDS with the standard Stable-Diffusion model to compute $L_{tex}^{ga}$. The results demonstrate a significant problem of geometry-texture misalignment. This issue arises because the standard Stable Diffusion model only utilizes text as a conditional input and lacks perception of geometry, therefore failing to provide geometry-aligned texture guidance.


\noindent\textbf{Texture prior SDS}. To assess the efficacy of our texture prior SDS, we compared it with two variants: one that solely relies on geometry-aware texture content SDS, denoted as \textit{\textbf{w/o}} $\boldsymbol{L_{tex}^{pr}}$, and another that excludes the use of $L_{tex}^{yuv}$, denoted as \textit{\textbf{w/o}} $\boldsymbol{L_{tex}^{yuv}}$. As shown in Columns 1,2 and 3 of Fig. \ref{fig:ablation-generation}(b), the results demonstrate that the \textit{\textbf{w/o}} $\boldsymbol{L_{tex}^{pr}}$ pipeline generates textures with significant noise and artifacts. The \textit{\textbf{w/o}} $\boldsymbol{L_{tex}^{yuv}}$ pipeline produces textures that generally adhere to the distribution of facial textures, but may exhibit brightness irregularities. The complete $L_{tex}^{pr}$ yields the best results.

\subsubsection{Effectiveness of SCP}

To evaluate the effectiveness of the proposed self-guided consistency preservation (SCP) in editing, we conduct the following ablation study. We make two variants: One variant, denoted as \textit{\textbf{w/o Reg}}, solely relies on $L_{edit}$ for editing without employing consistency regularization. The other variant, denoted as \textit{\textbf{w/o SC-weight}}, computes the consistency preservation regularization without using the self-guided consistency weight.

The results are shown in Fig. \ref{fig:ablation-scp}. While \textit{\textbf{w/o Reg}} shows noticeable editings following the instructions, unexpected alterations occur, such as the skin and hair of Scarlett turning purple, and Hulk's skin turning yellow. This inadequacy can be attributed to the absence of consistency constraints. On the other hand, \textit{\textbf{w/o SC-weight}} prevents undesirable changes in the results but hampers the effectiveness of editing, making it difficult to observe significant editing effects. In contrast, the full version of SCP achieves evident editing effects while preserving consistency in unaffected regions, thereby ensuring desirable editing outcomes.

\section{Conclusion}
\label{sec:conclusion}
We propose FaceG2E, a novel approach for generating diverse and high-quality 3D faces and performing facial editing using texts. With the proposed geometry-texture decoupled generation, high-fidelity facial geometry and texture can be produced. The designed self-guided consistency preserved 
editing enabling us to perform flexible editing, e.g., sequential editing. Extensive evaluations demonstrate that FaceG2E outperforms SOTA methods in 3D face synthesis.

Despite achieving new state-of-the-art results, we notice some limitations in FaceG2E. (1) The geometric representation restricts us from generating shapes beyond the facial skin, such as hair and accessories. (2) Sequential editing enables the synthesis of customized faces, but it also leads to a significant increase in time consumption. Each round of editing requires additional time.

{
    \small
    \bibliographystyle{ieeenat_fullname}
    \bibliography{main}
}

\clearpage

%
\begin{appendices}

\section{Appendix}
\subsection{Implementation Details}
\label{sec:1}
\noindent\textbf{Camera settings}.
During the optimization, We employ a camera with fixed intrinsic parameters: near=0.1, far=10, fov=12.59, rendering image size=224. For the camera extrinsics, we defined a set of optional viewing angles and randomly selected one of these angles as the rendering viewpoint for optimization in each iteration. The elevation angle $x \in {0, 10, 30}$, the azimuth angle $y \in \{0, 30, 60, 300, 330\}$, and the camera distance $d \in \{1.5, 3\}$. We set these extrinsics to ensure that the rendering always includes the facial region.

\noindent\textbf{Light settings}.
We utilize spherical harmonic (SH) to represent lighting. We pre-define 16 sets of spherical harmonic 3-band coefficients. In each iteration of rendering, we randomly select one set from these coefficients to represent the current lighting.

\noindent\textbf{Prompt engineering}.
In the generation stage, for the face description prompt of a celebrity or a character, we add the prefix `a zoomed out DSLR photo of '. We also utilize the view-dependent prompt enhancement. For the azimuth in (0,45) and (315,360), we add a suffix ` from the front view', for the azimuth in (45,135) and (225,315), we add a suffix ` from the side view'.

\noindent\textbf{SDS Time schedule}.
Following the Dreamfusion \cite{poole2022dreamfusion}, we set the range of $t$ to be between 0.98 and 0.02 in the SDS computation process. Besides, we utilize the linearly decreasing schedule for $t$, which is crucial for the stability of synthesis. As the iteration progresses from 0 to the final (e.g. iteration 400), our $t$ value linearly decreases from 0.98 to 0.02.

\subsection{User survey as ablation}
We conduct a user survey as ablation to further validate the effectiveness of our key design. A total of 100 volunteers participated in the experiment. We presented the results of our method and different degradation versions,  alongside the text prompts. Then we invited the volunteers to rate the facial generation and editing. The ratings ranged from 1 to 5, with higher scores indicating higher satisfaction. The user rating results are shown in Tab. \ref{table:1} and Tab. \ref{table:2}. The results indicate that removing any of our key designs during the face generation or face editing leads to a decrease in user ratings. This suggests that our key designs are necessary for synthesizing high-quality faces.

\begin{table}[h]
\centering

\resizebox{0.9\linewidth}{!}{
\centering

\begin{tabular}{cccc}
\cline{1-4}
\multicolumn{4}{c}{Generation}                                            \\ \cline{1-4}
\multicolumn{1}{c|}{ours} & w/o $L_{tex}^{yuv}$  & w/o $L_{tex}^{pr}$   & w/o $L_{tex}^{ga}$ \& $L_{tex}^{pr}$ \\ \cline{1-4}
\multicolumn{1}{c|}{3.82} & 3.77  & 2.59  & 1.78         \\ \cline{1-4}
\end{tabular}

}
\caption{Ablation study of face generation based on user ratings.}
\label{table:1}
\end{table}

\begin{table}[h]
\centering

\resizebox{0.6\linewidth}{!}{
\centering
\begin{tabular}{ccc}
\hline
\multicolumn{3}{c}{Editing}                           \\ \hline
\multicolumn{1}{c|}{ours} & w/o SC-weight  & w/o Reg     \\ \hline
\multicolumn{1}{c|}{3.95} & 2.55 & 2.28 \\ \hline
\end{tabular}
}
\caption{Ablation study of face editing based on user ratings.}
\label{table:2}
\end{table}

\subsection{More Relighting Results}
\begin{figure}[t]
    \centering
    \includegraphics[width=\linewidth]{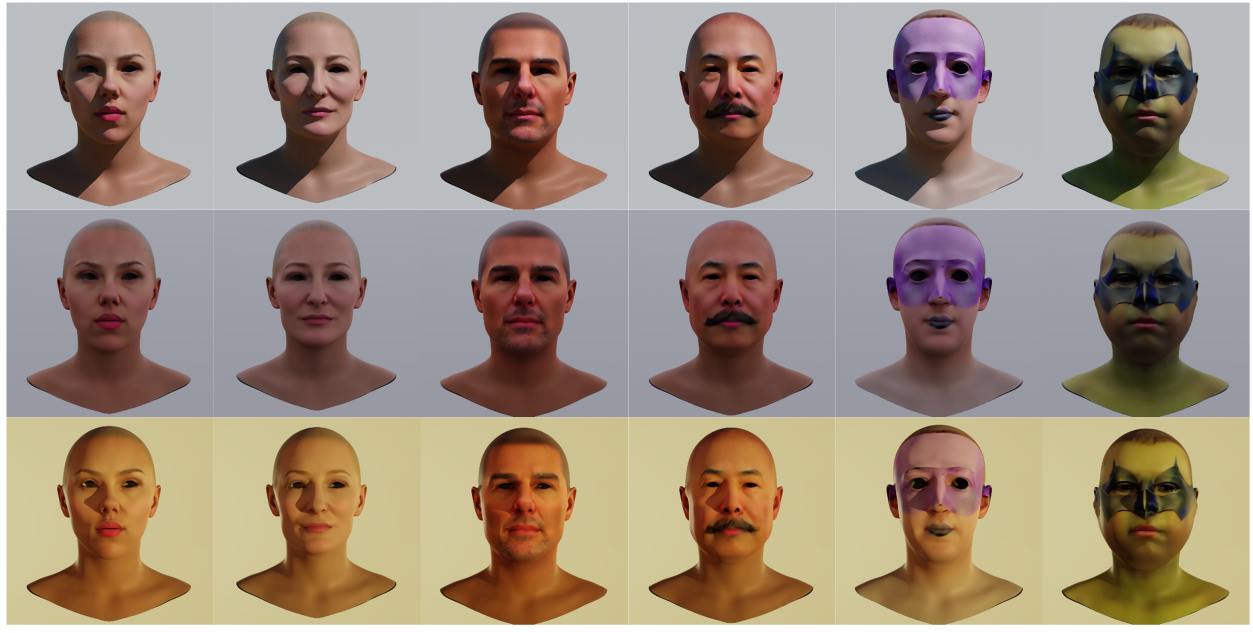}
    \captionsetup{width=\linewidth}
    \caption{Relighting of our synthesized 3D faces.}
    \label{fig:relighting}
\end{figure}

We present some more relighting results in Fig \ref{fig:relighting}. 
We recommend referring to the supplementary video or project page, where the video results can better demonstrate our animation and relighting effects.

\subsection{Generation with composed prompt}
Our sequential editing can synthesize complex 3D faces, an alternative approach is to combine all editing prompts into a composed prompt and generate the face in one step.

\begin{figure}[t]
    \centering
    \includegraphics[width=\linewidth]{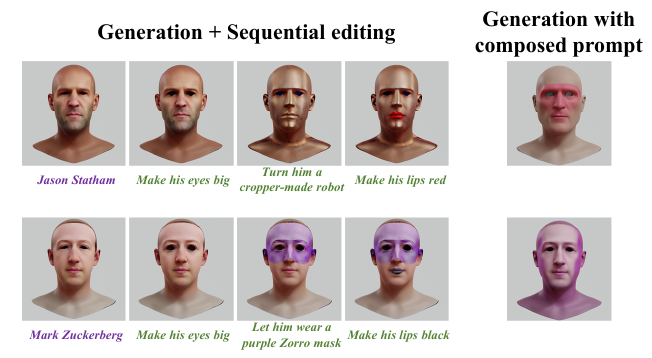}
    \captionsetup{width=\linewidth}
    \caption{Generation with composed prompt leads to the loss of concepts in prompts.}
    \label{fig:compose}
\end{figure}

In Fig.\ref{fig:compose}, we showcase the results generated from a composed prompt with our generation stage. It can be observed that directly generating with the composed prompt leads to the loss of certain concepts and details present in the prompts (e.g., the cropped-made effect in row 1, or the black lips in row 2). This underscores the necessity of the editing technique we propose for synthesizing customized faces.

\subsection{More Comparison Results}
We conduct more comparisons with more baseline methods. We add two baselines: a public implementation \cite{stabledreamfusion} for the Dreamfusion, and AvatarCraft \cite{jiang2023avatarcraft}, a SOTA text-to-3D avatar method that utilizes the implicit neutral field representation. We compare text-guided 3D face generation, single-round 3D face editing, and sequential 3D face editing. Note that baseline methods are not capable of directly editing 3D faces with text instruction (e.g., `make her old'), so we let them perform the editing by generating a face with the composed prompt. For example, `an old Emma Watson' is the composed prompt of `Emma Watson' and `Make her old'.

\begin{figure*}[t]
    \centering
    \includegraphics[width=0.9\linewidth]{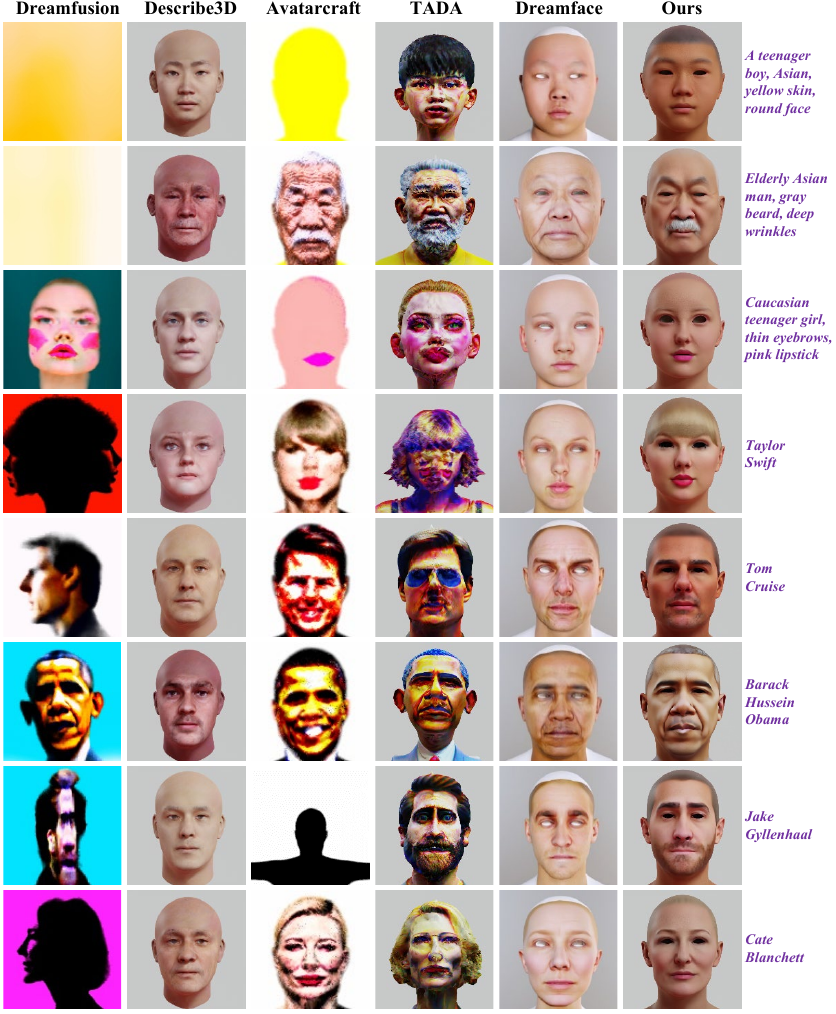}
    \captionsetup{width=\linewidth}
    \caption{Comparison on text-guided 3D face generation.}
    \label{fig:gen1}
\end{figure*}

\begin{figure*}[t]
    \centering
    \includegraphics[width=0.9\linewidth]{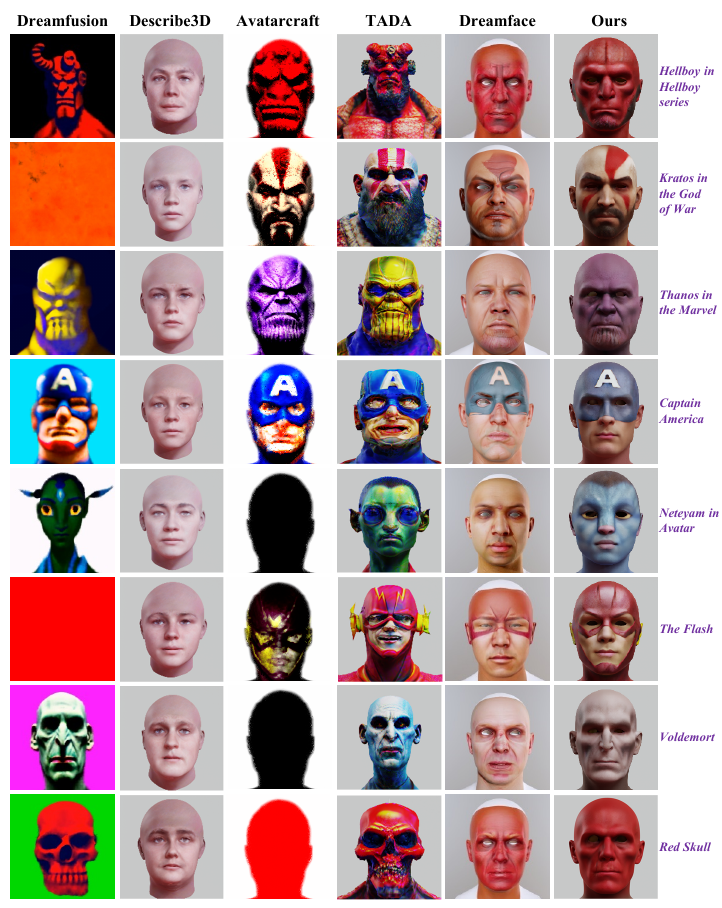}
    \captionsetup{width=\linewidth}
    \caption{Comparison on text-guided 3D face generation.}
    \label{fig:gen2}
\end{figure*}

\begin{figure*}[t]
    \centering
    \includegraphics[width=0.9\linewidth]{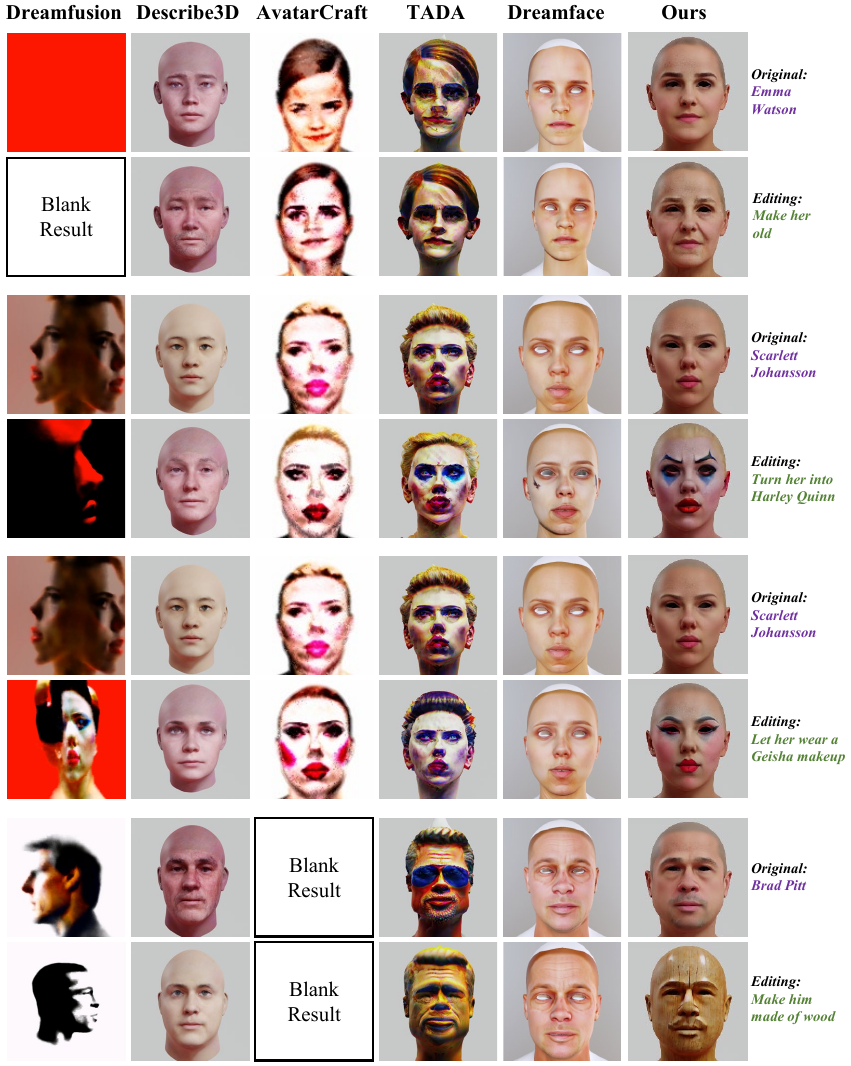}
    \captionsetup{width=\linewidth}
    \caption{Comparison on text-guided single-round 3D face editing.}
    \label{fig:edit1}
\end{figure*}

\begin{figure*}[t]
    \centering
    \includegraphics[width=0.9\linewidth]{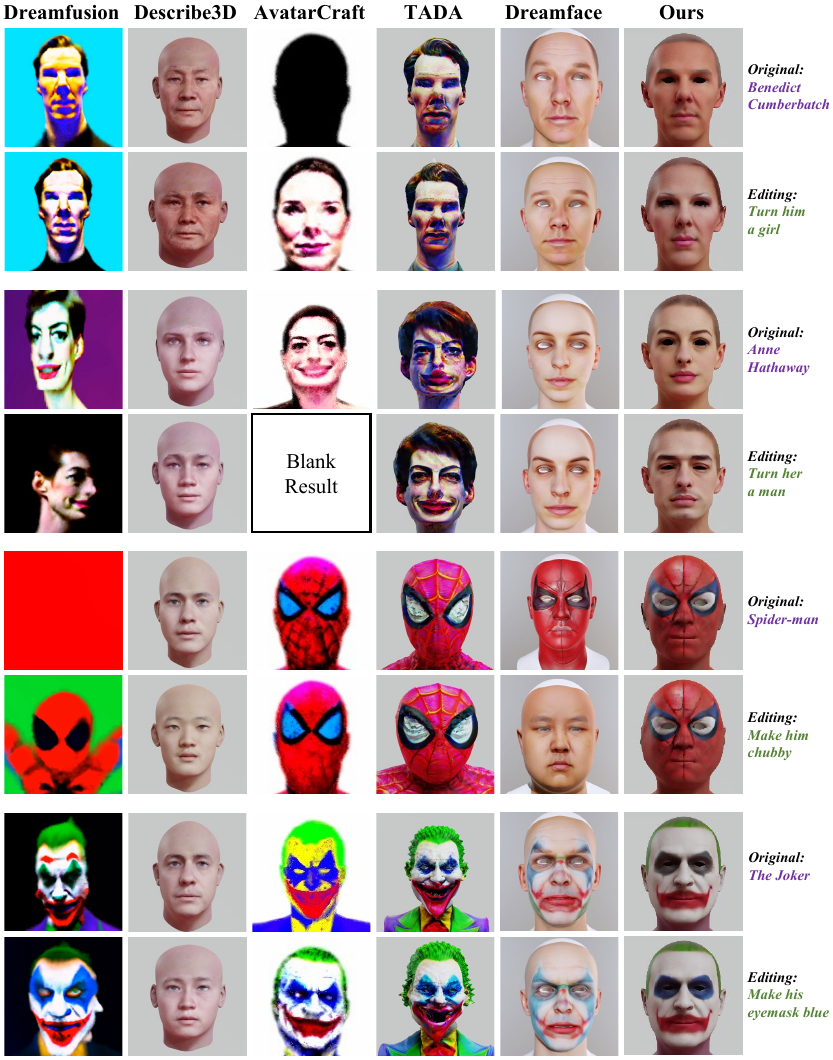}
    \captionsetup{width=\linewidth}
    \caption{Comparison on text-guided single-round 3D face editing.}
    \label{fig:edit2}
\end{figure*}

\begin{figure*}[t]
    \centering
    \includegraphics[width=0.9\linewidth]{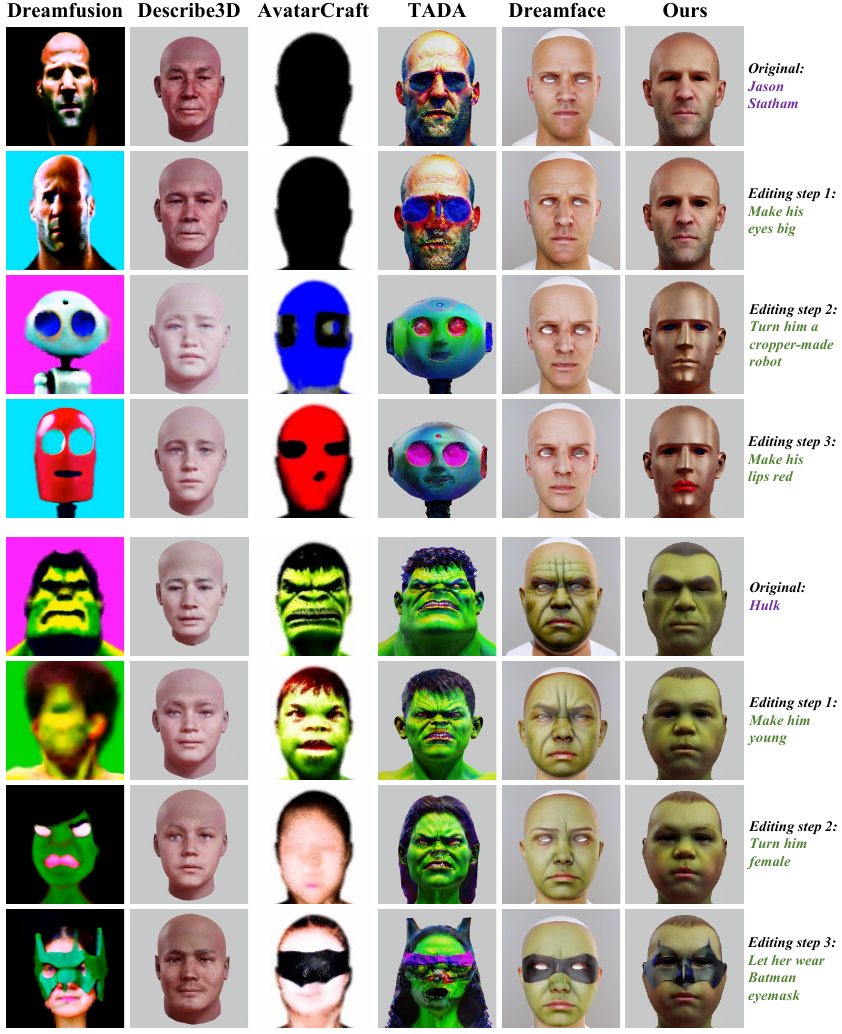}
    \captionsetup{width=\linewidth}
    \caption{Comparison on text-guided sequential 3D face editing.}
    \label{fig:seq12}
\end{figure*}

\begin{figure*}[t]
    \centering
    \includegraphics[width=0.9\linewidth]{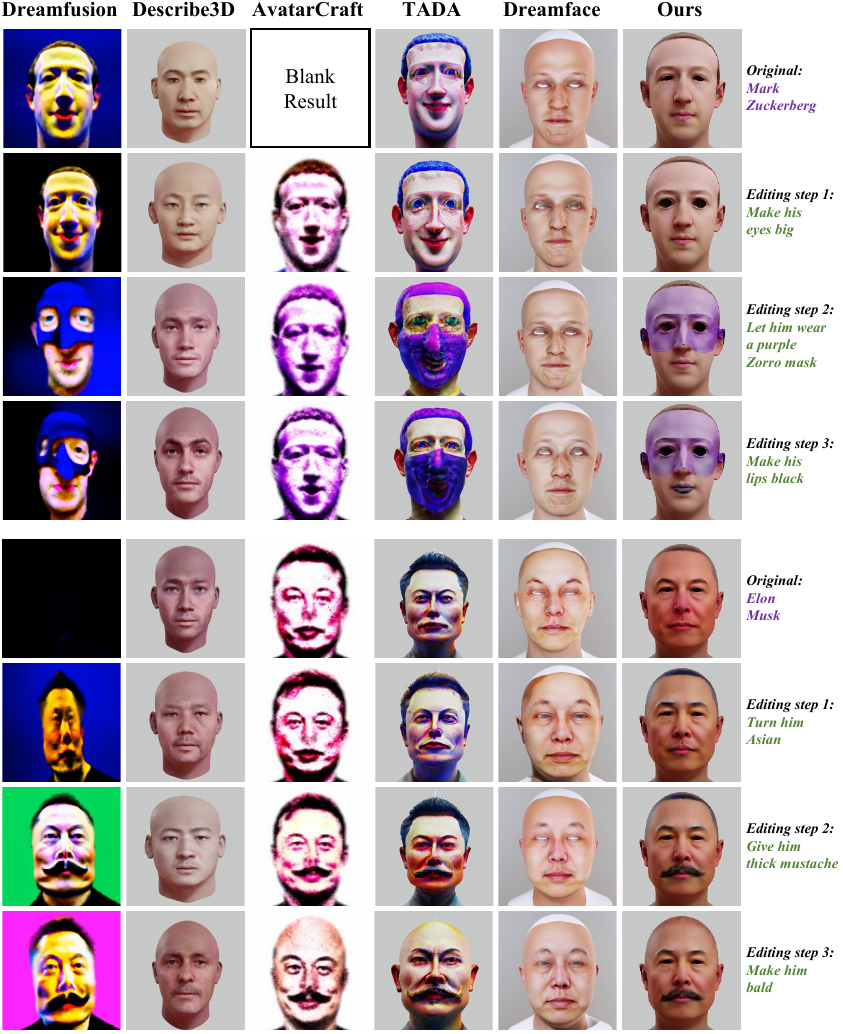}
    \captionsetup{width=\linewidth}
    \caption{Comparison on text-guided sequential 3D face editing.}
    \label{fig:seq34}
\end{figure*}

We present the 3D face generation results in Fig \ref{fig:gen1} and Fig \ref{fig:gen2}. The 3D face editing results are contained in Fig \ref{fig:edit1} and Fig \ref{fig:edit2}. The comparisons on sequential editing are presented in Fig \ref{fig:seq12} and Fig \ref{fig:seq34}. 
It should be noted that Dreamfusion \cite{stabledreamfusion} and Avatarcraft \cite{jiang2023avatarcraft} occasionally fail to produce meaningful 3D shapes and instead output a white background for some prompts. This issue could potentially be addressed by resetting the random seed, however, due to time constraints, we did not attempt repeated trials. We have labeled these examples as `Blank Result' in the figures.

\end{appendices}

\end{document}